\documentclass[10pt,twocolumn,letterpaper]{article}

\usepackage{iccv}
\usepackage{times}
\usepackage{epsfig}
\usepackage{graphicx}
\usepackage{amsmath}
\usepackage{amssymb}
\usepackage{makecell}
\usepackage[multiple]{footmisc}
\newcommand{\calL}{\mathcal{L}}
\newcommand{\calG}{\mathcal{G}}

\usepackage[breaklinks=true,bookmarks=false]{hyperref}

\iccvfinalcopy 

\def\iccvPaperID{****} 

\ificcvfinal\fi

\begin{document}

\title{SROBB: Targeted Perceptual Loss for Single Image Super-Resolution}

\author{Mohammad Saeed Rad$^{1}$\\
\and
Behzad Bozorgtabar$^{1}$\\ 
\and
Urs-Viktor Marti$^{2}$\\ 
\and
Max Basler$^{2}$\\ 
\and
Haz{\i}m Kemal Ekenel$^{1,\, 3}$\\ 
\and
Jean-Philippe Thiran$^{1}$\\
\and
$^{1}$LTS5, EPFL, Switzerland \quad \quad
$^{2}$AI Lab, Swisscom AG, Switzerland \quad \quad
$^{3}$SiMiT Lab, ITU, Turkey\\
{\tt\small \{saeed.rad, firstname.lastname\}@epfl.ch \qquad \{firstname.lastname\}@swisscom.com}
}

\maketitle

\ificcvfinal\thispagestyle{empty}\fi

\begin{abstract}
By benefiting from perceptual losses, recent studies have improved significantly the performance of the super-resolution task, where a high-resolution image is resolved from its low-resolution counterpart. Although such objective functions generate near-photorealistic results, their capability is limited, since they estimate the reconstruction error for an entire image in the same way, without considering any semantic information. In this paper, we propose a novel method to benefit from perceptual loss in a more objective way. We optimize a deep network-based decoder with a targeted objective function that penalizes images at different semantic levels using the corresponding terms. In particular, the proposed method leverages our proposed OBB (Object, Background and Boundary) labels, generated from segmentation labels, to estimate a suitable perceptual loss for boundaries, while considering texture similarity for backgrounds. We show that our proposed approach results in more realistic textures and sharper edges, and outperforms other state-of-the-art algorithms in terms of both qualitative results on standard benchmarks and results of extensive user studies.
\end{abstract}

\section{Introduction}
\label{sec:intro}

Single image super-resolution (SISR) aims at solving the problem of recovering a high-resolution (HR) image from its low-resolution (LR) counterpart. SISR is a classic ill-posed problem that has been one of the most active research areas since the work of Tsai and Huang~\cite{paper_tsai} in 1984. In recent years, this problem has been revolutionized by the significant advances in convolutional neural networks (CNNs) and has resulted in better reconstructions of high-resolution pictures than classical approaches~\cite{paper_cnn_sr, paper_srcnn, paper_DRCN}. More recently, another breakthrough has been made in SISR by employing perceptual loss functions for training feed-forward networks, instead of using per-pixel loss functions, e.g., mean squared error (MSE) ~\cite{paper_perc_0,paper_enhanced,paper_twitter_0}. It tackled the problem of blurred textures caused by optimization of MSE, and alongside with adversarial loss~\cite{paper_gan}, it resulted in near-photorealistic reconstruction in terms of perceived image quality.  

\begin{figure}[t]
\begin{center}
   \includegraphics[width=1.01\linewidth]{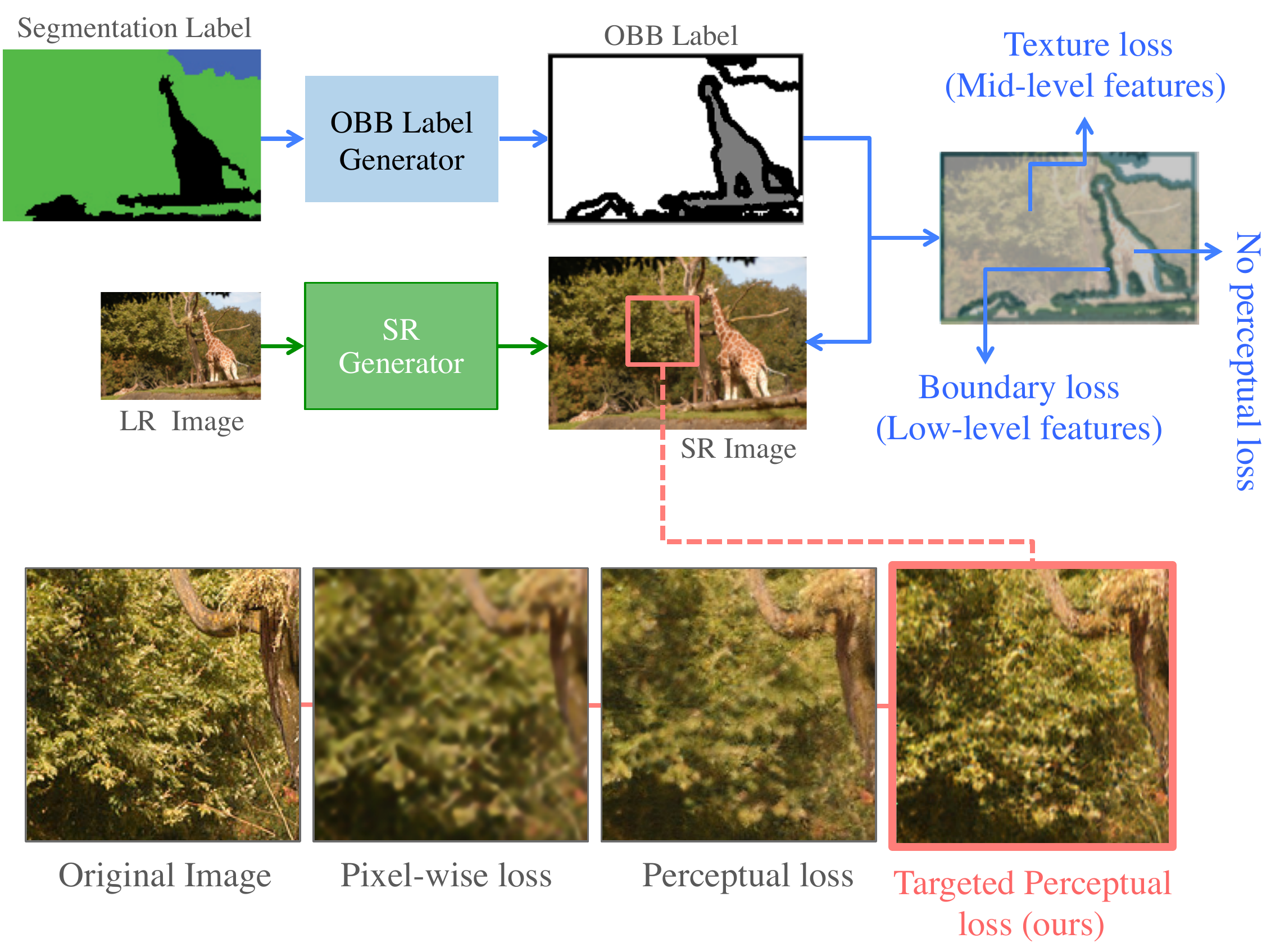}
\end{center}
   \caption{We propose a method for exploiting the segmentation labels during training to resolve a high resolution image at different semantic levels considering their characteristics; we optimize our SISR model by minimizing perceptual errors that correspond to edges only at object boundaries and the texture on the background area, respectively. Results from left to right: original image, super-resolved images using only pixel-wise loss function, pixel-wise loss + perceptual loss function and pixel-wise loss + targeted perceptual loss function (ours), respectively.}
\label{fig:intro}
\end{figure}

\cite{paper_enhanced} and \cite{paper_twitter_0} benefit from the idea of using perceptual similarity as a loss function; they optimize their models by comparing the ground-truth and the predicted super-resolved image (SR) in a deep feature domain by mapping both HR and SR images into a feature space using a pre-trained classification network. Although this similarity measure in feature space, namely the perceptual loss, has shown a great success in SISR, applying it as it is on a whole image, without considering the semantic information, limits its capability.

To better understand this limitation, let us have a brief overview of the perceptual loss and see what a pre-trained classification network optimizes; considering a pre-trained CNN, in an early convolutional layer, each neuron has a receptive field with the size and shape of the inputs that affects its output. Small kernels, which are commonly used by state-of-the-art approaches, have also small receptive fields. As a result, they can only extract low-level spatial information. Intuitively, each neuron captures relations between nearby inputs considering their local spatial relations. These local relations are mostly presenting information about edges and blobs. As we proceed deeper in the network, the receptive field of each neuron with respect to earlier layers becomes larger. Therefore, deep layers start to learn features with global semantic meanings and abstract object information, and less fine-grained spatial details, while still using small kernels. This fact has also been shown by \cite{paper_visualization_1,paper_visualization_2}, where they used some visualization techniques and investigated the internal working mechanism of the VGG network~\cite{paper_vgg} by visualization of the information kept in each CNN layer. 

Regarding the perceptual function, state-of-the-art approaches use different levels of features to restore the original image; this choice determines whether they focus on local information such as edges, mid-level features such as textures or high-level features corresponding to semantic information. In these works, perceptual loss has been calculated for an entire image in the same way, meaning that the same level of features has been used either on edges, foreground or on the image background. For example, minimizing the loss for details of the edges inside a random texture, such as the texture of a tree, would force the network to consider an unnecessary penalty and learn less informative features; the texture of a tree could still be realistic in the SR image without having close edges to the HR image. On the other hand, minimizing the loss by using mid-level features (more appropriate for the textures) around edges would not intuitively create sharper edges and would only introduce ``noisy'' losses.

To address the above issue, we propose a novel method to benefit from perceptual loss in a more objective way. Figure \ref{fig:intro} shows an overview of our proposed approach. In particular, we use pixel-wise segmentation annotations to build our proposed OBB labels to be able to find targeted perceptual features that can be used to minimize appropriate losses to different image areas: e.g., edge loss for edges and textures' loss for image textures during  training. We show that our approach using targeted perceptual loss outperforms other state-of-the-art algorithms in terms of both qualitative results and user study experiments, and result in more realistic textures and sharper edges.

\section{Related work}
In this section, we review relevant CNN-based SISR approaches. This field has witnessed a variety of end-to-end deep network architectures: \cite{paper_DRCN} formulated a recursive CNN and showed how deeper network architectures increase the performance of SISR. \cite{paper_twitter_0,paper_enhanced,Zhang_2018_CVPR} used the concept of residual blocks~\cite{HeZRS15} and skip-connections \cite{HeZR016,paper_DRCN} to facilitate the training of CNN-based decoders. \cite{paper_edsr} improved their models by expanding the model size. \cite{paper_esrgan} removed batch normalization in conventional residual networks and used several skip connections to improve the results of seminal work of \cite{paper_twitter_0}. Laplacian pyramid structure~\cite{paper_lapsrn} has been proposed to progressively reconstruct the sub-band residuals of high-resolution images. \cite{memnet} proposed a densely connected network that uses a memory block consisting of a recursive unit and a gate unit, to explicitly mine persistent memory through an adaptive learning process. \cite{paper_rcan} proposed a channel attention mechanism to adaptively rescale channel-wise features by considering the inter-dependencies among channels. Besides supervised learning, other methods like unsupervised learning \cite{unsupervisedSR} and reinforcement learning \cite{crafting} were also introduced to solve the SR problem. 

Despite variant architectures proposed for the SISR task, the behavior of optimization-based methods is principally driven by the choice of the objective function. The objective functions used by these works mostly contain a loss term with the pixel-wise distance between the super-resolved and the ground-truth HR images. However, using this function alone leads to blurry and over-smoothed super-resolved images due to the pixel-wise average of all plausible solutions.

Perceptual-driven approaches added a remarkable improvement to image super-resolution in terms of the visual quality. Based on the idea of perceptual similarity \cite{bruna2015super}, perceptual loss \cite{paper_perc_0} is proposed to minimize the error in a feature space using specific layers of a pre-trained feature extractor, for example VGG \cite{paper_vgg}. A number of recent papers have used this optimization to generate images depending on high-level extracted features \cite{GatysEB15,GatysEB15a,YosinskiCNFL15,SimonyanVZ13, paper_pirm_w2}. In a similar work, contextual loss \cite{mechrez2018learning} is proposed to generate images with natural image statistics, which focuses on the feature distribution rather than merely comparing the appearance. \cite{paper_twitter_0} proposed to use adversarial loss in addition to the perceptual loss to favor outputs residing on the manifold of natural images. The SR method in \cite{paper_enhanced} develops a similar approach and further explores a patch-based texture loss. Although these works generate near-photorealistic results, they estimate the reconstruction error for an entire image in the same way, without benefiting from any semantic information that could improve the visual quality.

Many studies such as \cite{5701777, 5540206, eth_biwi_01235} also benefit from prior information for SISR. Most recently, \cite{paper_segmentation} used an additional segmentation network to estimate probability maps as prior knowledge and used them in the existing super-resolution networks. Their segmentation network is pre-trained on the COCO dataset \cite{lin2014microsoft} and then is fine-tuned on the ADE dataset \cite{zhou2017scene}. Their approach recovers more realistic textures faithful to categorical priors; however, it requires a segmentation map at test-time. \cite{saeed_SRSEG} addressed this issue by proposing a method based on multitask learning simultaneously for SR and semantic segmentation tasks.

In this work, we investigate a novel way to exploit semantic information within an image, yielding photo-realistic super-resolved images with fine-structures.

\section{Methodology}
 
Following recent approaches \cite{paper_twitter_0, paper_segmentation, paper_VSR} for image and video super-resolution, we benefit from deep networks with residual blocks to build-up our decoder. As explained previously, in this paper, we focus on the definition of the objective function used to train our network; we introduce a loss function containing three terms: 1- Pixel-wise loss (MSE), 2- adversarial loss, and 3- our novel targeted perceptual loss function. The MSE and adversarial loss terms are defined as follows:

\begin{itemize}
  \item \textbf{Pixel-wise loss} It is by far the most commonly used loss function in SR. It calculates the pixel-wise mean squared error (MSE) between the original image and the super-resolved image in the image domain \cite{paper_enhanced,paper_srcnn,paper_deep_sr_1}. The main drawback of using it as a stand-alone objective function is mostly resolving an over-smoothed reconstruction. The network trained with the MSE loss seeks to find pixel-wise averages of plausible solutions, which results in poor perceptual qualities and lack of high-frequency details in the edges and textures.
  \item \textbf{Adversarial loss} Inspired by \cite{paper_twitter_0}, we formulate our SR model in an adversarial setting, which provides a feasible solution. In particular, we use an additional network (discriminator) that is alternatively trained to compete with our SR decoder. The generator (SR decoder) tries to generate fake images to fool the discriminator, while the discriminator aims at distinguishing the generated results from real HR images. This setting results in perceptually superior solutions to the ones obtained by minimizing pixel-wise MSE and classic perceptual losses. The discriminator used in this work is defined in more details in Section \ref{subsec:architecture}.

\end{itemize}

Our proposed targeted perceptual loss is described in the following subsection.
\subsection{Targeted perceptual loss}

The state-of-the-art approaches such as \cite{paper_enhanced} and \cite{paper_twitter_0} estimate perceptual similarity by comparing the ground-truth and the predicted super-resolved image in a deep feature domain by mapping both HR and SR images into a feature space using a pre-trained classification network, e.g., VGG~\cite{paper_vgg}. The output of a specific convolutional layer is used as the feature map. These approaches usually minimize the $l_{2}$ distance of the feature maps. In order to understand why minimizing this loss term in combination with adversarial and MSE losses is effective and results in more photorealistic images, we investigate the nature of the CNN layers used for the perceptual loss. Then, we propose a novel approach to take advantage of the perceptual similarity in a targeted manner and reconstruct more appealing edges and textures.

\begin{figure}[t]
\begin{center}
   \includegraphics[height=4.0cm]{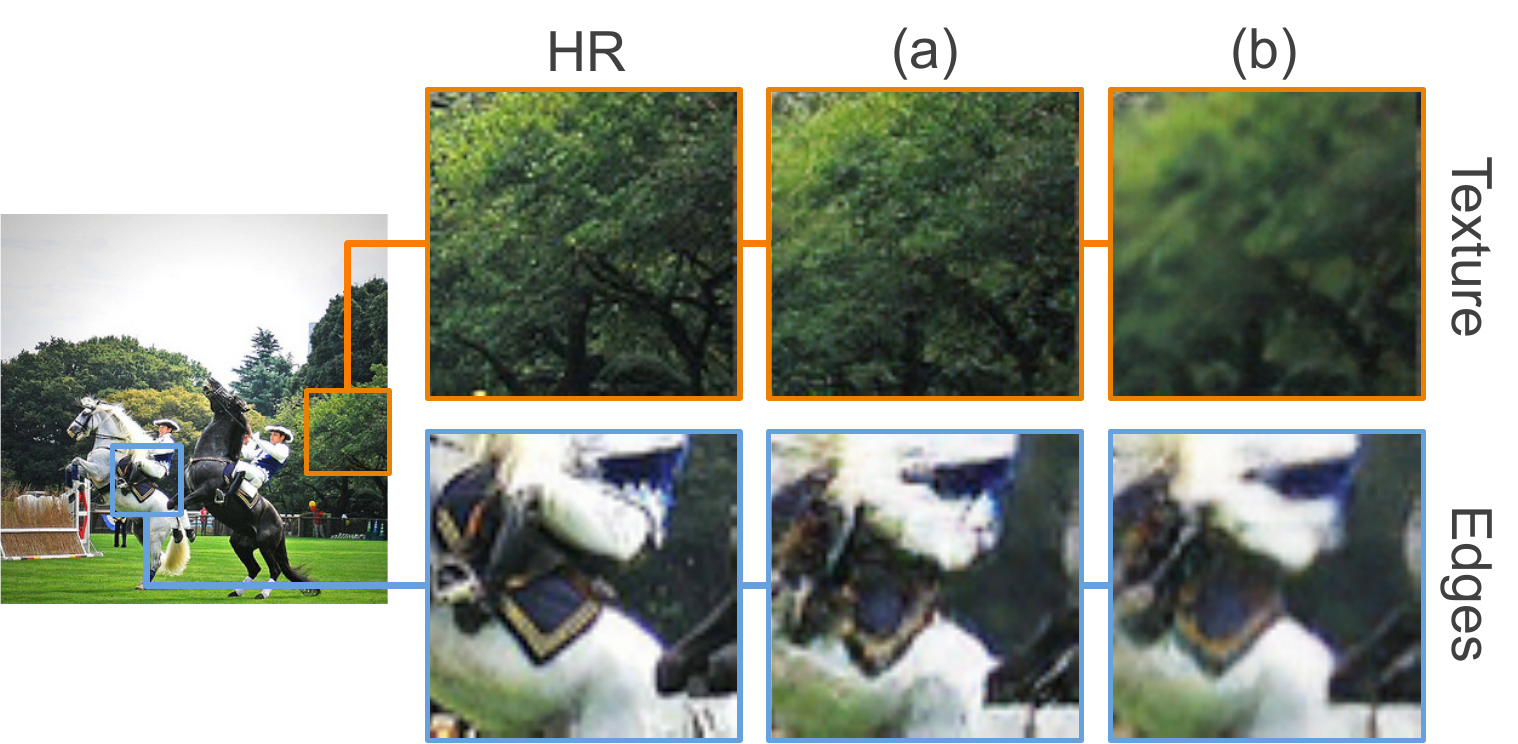}
\end{center}
   \caption{The effect of choosing different CNN layers to estimate the perceptual loss on different regions of an image, e.g., edges and textures: (a) using a deeper convolutional layer (mid-level features), ReLU~4-1 of VGG-16~\cite{paper_vgg} and, (b) using an early convolutional layer (low-level features), ReLU~1-2 of the VGG-16 network.}
\label{fig:perceptual_loss}
\end{figure}

As explained previously, early layers of a CNN return low-level spatial information regarding local relations, such as information about edges and blobs. As we proceed towards deeper layers, we start to learn higher level features with more semantic meaning and abstract object information, and less fine-grained spatial details from an image. In this fashion, mid-level features are mostly representing textures and high-level features amount to the global semantic meaning. Figure~\ref{fig:perceptual_loss} shows the difference between shallow and deep layers of a feature extractor, the VGG-16 in our case; two different layers, ReLU~1-2 and ReLU~4-1, are used to compute the perceptual loss and reconstruct an image. We compare each case on an edge and a texture region. In this figure, we can see using low-level features is more effective for reconstructing edges, while mid-level features resolve closer textures to the original image.

The targeted loss function tries to favor more realistic textures around areas, where the type of the textures seems to be important, e.g., a tree, while trying to resolve sharper edges around boundary area. To do so, we first define three types of regions in an image: 1- background, 2- boundaries, and 3- objects, then, we compute the targeted perceptual loss for each region using a different function.

\begin{itemize}
  \item \textbf{Background} ($\calG_{b}$) We consider four classes as background: ``sky'', ``plant'', ``ground'' and ``water''. We chose these categories because of their specific appearance; the overall texture in areas with these labels are more important than local spatial relations and edges. We compute mid-level CNN features to estimate the perceptual similarity between SR and HR images. Here, we use the ReLU 4-3 layer of the VGG-16 for this purpose.
  \item \textbf{Boundary} ($\calG_{e}$) All edges separating objects and the background are considered as boundaries. With some pre-processing (explained in more detail in Section~\ref{subsec:obb}), we broaden these edges to have a strip passing through all boundaries. We estimate the feature distance of an early CNN layer between SR and HR images, which focuses more on low-level spatial information, mainly edges and blobs. In particular, we minimize the perceptual loss at the ReLU 2-2 layer of the VGG-16.
  \item \textbf{Object} ($\calG_{o}$) Because of the huge variety of objects in the real world in terms of shapes and textures, it is challenging to decide whether it is more appropriate to use features from early or deeper layers for the perceptual loss function; for example, in an image of a zebra, sharper edges are more important than the overall texture. Having said that, forcing the network to estimate the precise edges in a tree could mislead the optimization procedure. Therefore, we do not consider any type of perceptual loss on areas defined as objects by weighting them to zero and rely on the MSE and adversarial losses. However, intuitively, resolving more realistic textures and sharper edges by the ``background'' and ``boundary'' perceptual loss terms would result in more appealing objects, as well. 
\end{itemize}

To compute the perceptual loss for a specific image region, we make binary segmentation masks of the semantic classes (having a pixel value of 1 for the class of interest and 0 elsewhere). Each mask categorically represents a different region of an image and is element-wise multiplied by the HR image and the estimated super-resolved image SR, respectively. In other words, for a given category, the image is converted to a black image with only one visible area on it, before being passed through the CNN feature extractor. Masking an image in this way creates also new artificial boundaries between black regions and the visible class. As a consequence, extracted features contain information about the artificial edges which do not exist in a real image. As the same mask is applied on both HR and the reconstructed image, the feature distance between these artificial edges will be close to zero and it does not affect the total perceptual loss. We can conclude that all non-zero distances in feature space between the masked HR and super-resolved image are corresponds to the contents of the visible area of that image: corresponds to edges by using a mask for boundaries ($M_{OBB}^{boundaries}$) and corresponds to textures by using a mask for the background ($M_{OBB}^{background}$).

The overall targeted perceptual loss function is given as:

\begin{align}\label{eq:targeted_loss}
\calL_{perc.} &= \alpha \cdot \calG_{e}( I^{SR}\circ M_{OBB}^{boundary}, I^{HR}\circ M_{OBB}^{boundary}) \nonumber \\
&\quad  + \beta \cdot \calG_{b}( I^{SR}\circ M_{OBB}^{background}, I^{HR}\circ M_{OBB}^{background})\nonumber \\
&\quad + \gamma \cdot \calG_{o}
\end{align} where $\alpha$, $\beta$ and $\gamma$ are the corresponding weights of the loss terms used for the boundary, background, and object, respectively. $\calG_{e} (\cdot)$, $\calG_{b} (\cdot)$ and $\calG_{o} (\cdot)$ are the functions to calculate feature space distances between any two given images for the boundaries, background, and objects, respectively. In this equation, $\circ$ denotes element-wise multiplication. As discussed earlier, we do not consider any perceptual loss for objects areas, therefore, we set $\gamma$ directly to zero. The value of other weights are discussed in detail in Section~\ref{subsec:parameters}.

In the following subsection, we describe how to build a label indicating objects, the background, and boundaries for the training images. This labeling approach helps us to use specific masks for each class of interest ($M_{OBB}^{object}$, $M_{OBB}^{background}$ and $M_{OBB}^{boundary}$) and to guide our proposed perceptual losses to focus on area of interest within the image.

\begin{figure}[h]
\begin{center}
   \includegraphics[height=3.0cm]{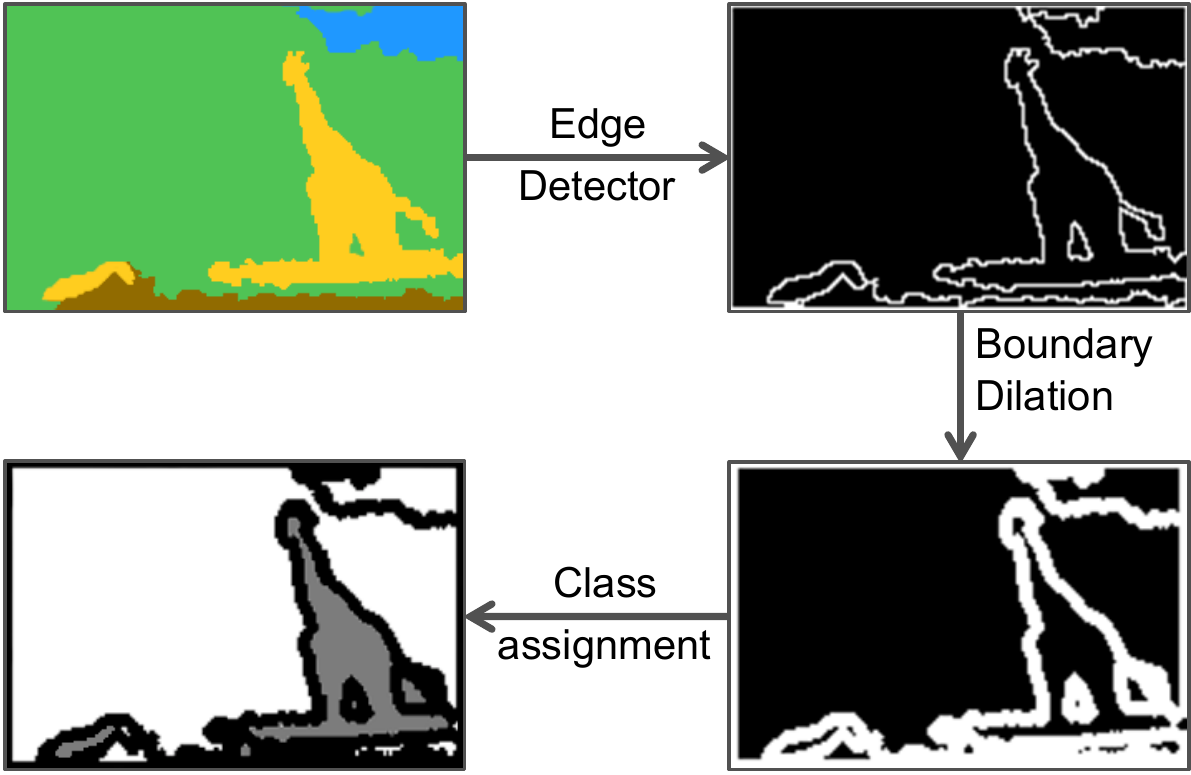}
\end{center}
   \caption{Constructing an OBB label. We assign each area to one of the ``Object'', ``Background'' or ``Boundary'' classes based on their initial pixel-wise labels.}
\label{fig:obb_label}
\end{figure}
\subsection{OBB: Object, background and boundary label}
\label{subsec:obb}
In order to make full use of the perceptual loss-based image super-resolution, we enforce semantic details (where objects, the background, and boundaries appear on the image) via our proposed targeted loss function. In addition, existing annotations for the segmentation task, e.g., \cite{paper_coco_stuff} only provide spatial information about objects and the background, and they do not use classes representing the edge areas, namely boundaries in this paper. Therefore, inspired by \cite{saeed_SRSEG}, we propose our labeling approach (Figure~\ref{fig:obb_label}) to provide a better spatial control of the semantic information for the images.
\begin{figure*}[ht]
\vspace{-4mm} 
\begin{center}
   \includegraphics[height=5.25cm]{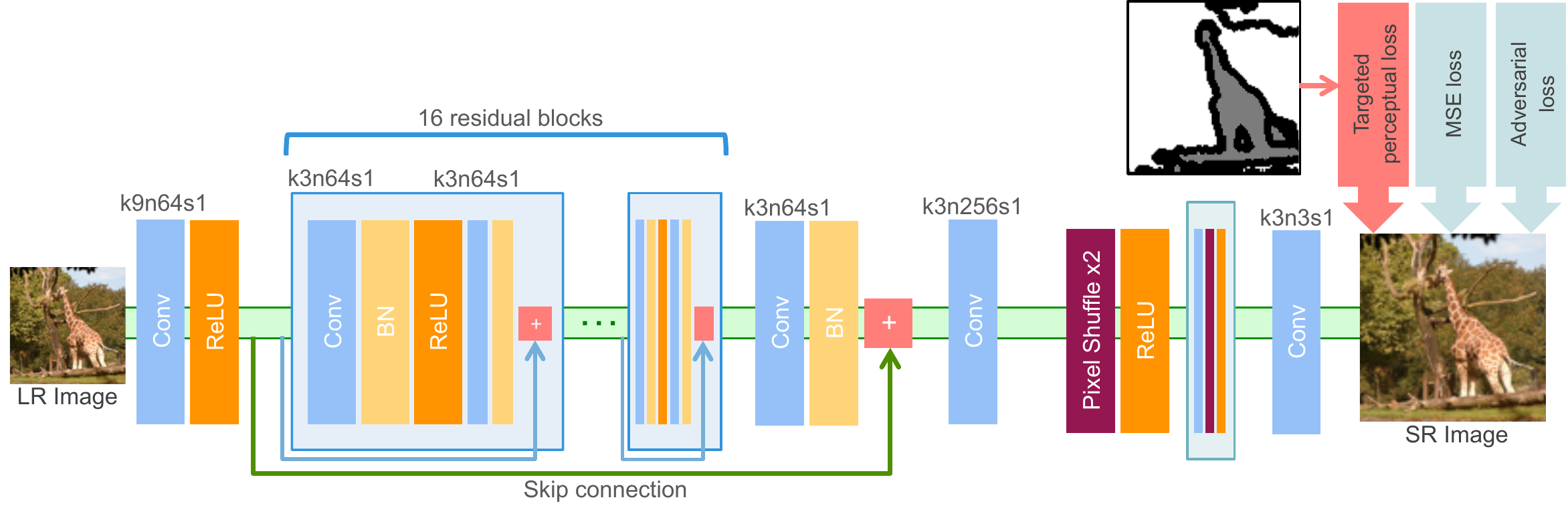}
\end{center}
   \caption{Schematic diagram of the SR decoder. We train the SR decoder using the targeted perceptual loss alongside with MSE and adversarial losses. In this schema, $k$, $n$ and $s$ correspond to kernel size, number of feature maps and stride size, respectively.}
\label{fig:archi}
\end{figure*}

To create such labels (OBB label), first, we calculate the derivative of the segmentation label in the color-space to estimate the edges between object classes in the segmentation label as well as the edges between objects and background of the image. In order to have a thicker strip around all edges separating different classes, we compute the dilation with a disk of size $d_1$. We label the resulted area as ``boundary'' class, which covers boundaries between different classes inside an image. In particular, we consider ``sky'', ``plant'', ``ground'', and ``water'' classes from the segmentation labels as the ``Background''. All remaining object classes are considered as the ``object'' class.


\subsection{Architecture}
\label{subsec:architecture}
For a fair comparison with the SRGAN method ~\cite{paper_twitter_0} and performing an ablation study of the proposed targeted perceptual loss, we use the same SR decoder as the SRGAN. The generator network is a feed-forward CNN. The input image $I^{LR}$ is passed through a convolution block followed by a ReLU activation layer. The output is subsequently passed through 16 residual blocks with skip connections. Each block has two convolutional layers with $3\times3$ filters and $64$ channels feature maps, each one followed by a batch normalization and ReLU activation. The output of the final residual block is concatenated with the features of the first convolutional layer and is then passed through two upsampling blocks, where each one doubles the size of the feature map. Finally, the result is filtered by a last convolution layer to get the super-resolved image $I^{SR}$. In this paper, we use a scale factor of four; depending on the desired scaling factor, the number of upsampling blocks could be modified. An overview of the architecture is shown in Figure \ref{fig:archi}.

The discriminator network consists of multiple convolutional layers with an increasing number of channels of the feature maps by a factor of $2$, from $64$ to $512$. We use Leaky-ReLU and strided convolutions to reduce the image dimension while doubling the number of features. The resulting $512$ feature maps are passed through two dense layers. Finally, the discriminator network classifies the image as real or fake by the final sigmoid activation function.

\section{Experimental Results}
In this section, first, we describe the training parameters and dataset in details, then we evaluate our proposed method in terms of qualitative, quantitative, and running costs analysis.

\subsection{Dataset and parameters}
\label{subsec:parameters}
To create OBB labels, we use a random set of 50K images from the COCO-Stuff dataset \cite{paper_coco_stuff}, which contains semantic labels of 91 classes for the segmentation task. In this paper, we considered landscapes with one or more of the ``Sky'', ``Plant'', ``Ground'', and ``Water'' classes. We group these classes into one ``Background'' class. We use our proposed technique in Section~\ref{subsec:obb} to convert pixel-wise segmentation annotations to OBB labels. In order to obtain LR images, we use the MATLAB imresize function with the bicubic kernel and the anti-aliasing filter. All experiments were performed with a downsampling factor of four. 

The training process was done in two steps; first, the SR decoder was pre-trained for 25 epochs with only pixel-wise mean squared error as the loss function. Then the proposed targeted perceptual loss function, as well as the adversarial loss were added and the training continued for 55 more epochs. The weights of each term in the new targeted perceptual loss, $\alpha$ and $\beta$, were set to $2 \times 10^{-6}$ and $1.5 \times 10^{-6}$, respectively. The weights of adversarial and MSE loss function, as in~\cite{paper_twitter_0}, were set to $1.0$ and $1 \times 10^{-3}$, respectively. We set $d1$, the diameter of the disk used to generate OBB labels, to $2.0$. The Adam optimizer \cite{paper_adam} was used during both steps. The learning rate was set to $1 \times 10^{-3}$ and then decayed by a factor of 10 every 20 epochs. We also alternately optimized the discriminator with similar parameters to those proposed by \cite{paper_twitter_0}.

\begin{figure*}[h]
\vspace{-2mm}
\begin{center}
   \includegraphics[width=0.96\linewidth]{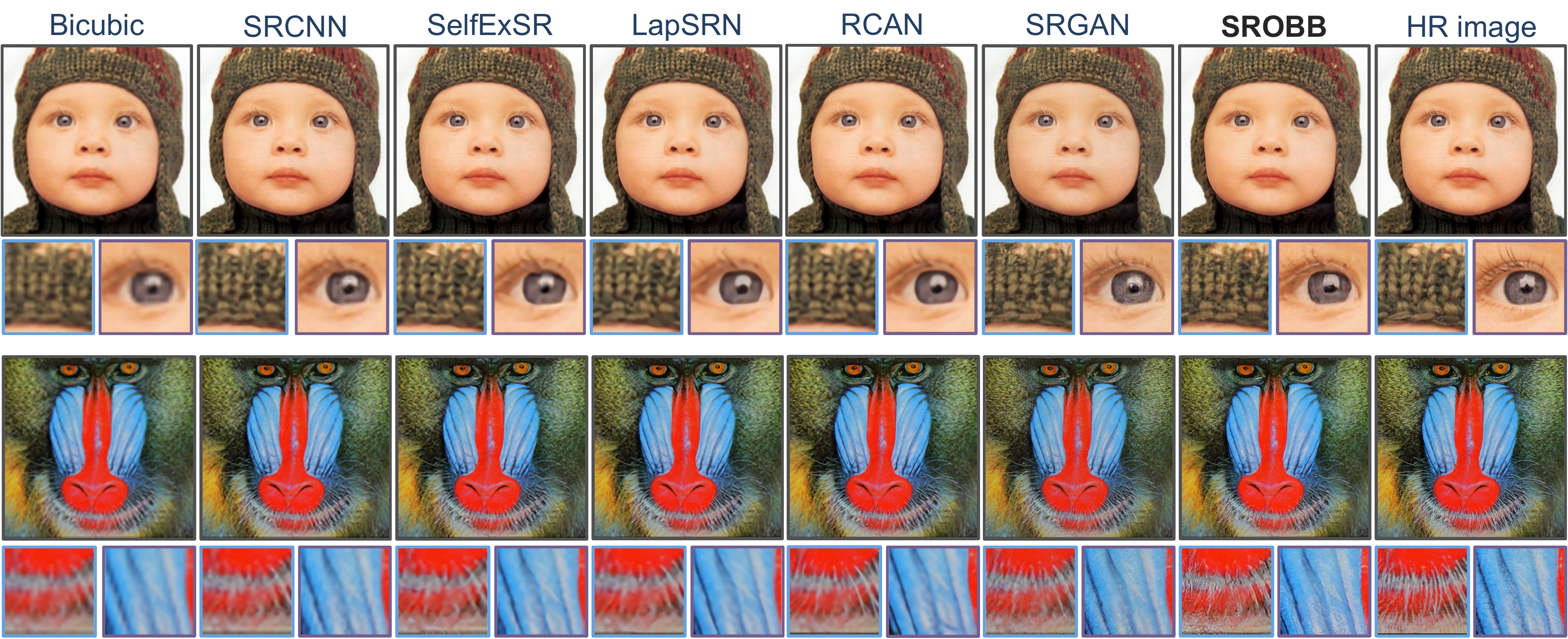}
\end{center}
   \caption{Sample results on the ``baby'' (top) and ``baboon'' (bottom) images from Set5~\cite{paper_set5} and Set14~\ref{fig:set5set14} datasets, respectively. From left to right: bicubic, SRCNN~\cite{paper_srcnn}, SelfExSR~\cite{paper_selfexsr}, LapSRN~\cite{paper_lapsrn}, RCAN~\cite{paper_rcan}, SRGAN~\cite{paper_twitter_0} and SROBB (ours), HR image, respectively.}
\label{fig:set5set14}
\vspace{-2mm}
\end{figure*}

\subsection{Qualitative Results}
\subsubsection{Results on Set5 and Set14}
Our approach focuses on optimizing the decoder with perceptual loss terms targeting boundaries and background by exploiting segmentation labels. Although, we do not apply the perceptual losses specifically on objects regions, our experiment shows that the trained model generalized in a way that it reconstructs more realistic objects compared to other approaches. We evaluate the quality of object reconstruction by performing qualitative experiments on two widely used benchmark datasets: Set5~\cite{paper_set5} and Set14~\cite{paper_set14}, where unlike our training set, in most of the images, outdoor background scenes are not present. Figure~\ref{fig:set5set14} compares the results of our SR model on the ``baby'' and ``baboon'' images and the recent state-of-the-art methods including: bicubic, SRCNN~\cite{paper_srcnn}, SelfExSR~\cite{paper_selfexsr}, LapSRN~\cite{paper_lapsrn}, RCAN~\cite{paper_rcan} and SRGAN~\cite{paper_twitter_0}. In the ``baboon'' image, we could generate more photo-realistic images with sharper edges compared to other methods while having competitive results for the ``baby'' image with SRGAN. Their results were obtained by using their online supplementary materials \footnote{\url{https://github.com/jbhuang0604/SelfExSR}} \footnote{\url{https://github.com/phoenix104104/LapSRN}} \footnote{\url{https://twitter.app.box.com/s/lcue6vlrd01ljkdtdkhmfvk7vtjhetog}}. More qualitative results of Set5 and Set14 images are provided in the supplementary material.

\subsubsection{Results on the COCO-Stuff dataset}
\begin{figure*}
\vspace{-3mm}
\begin{center}
\includegraphics[width=1.0\linewidth]{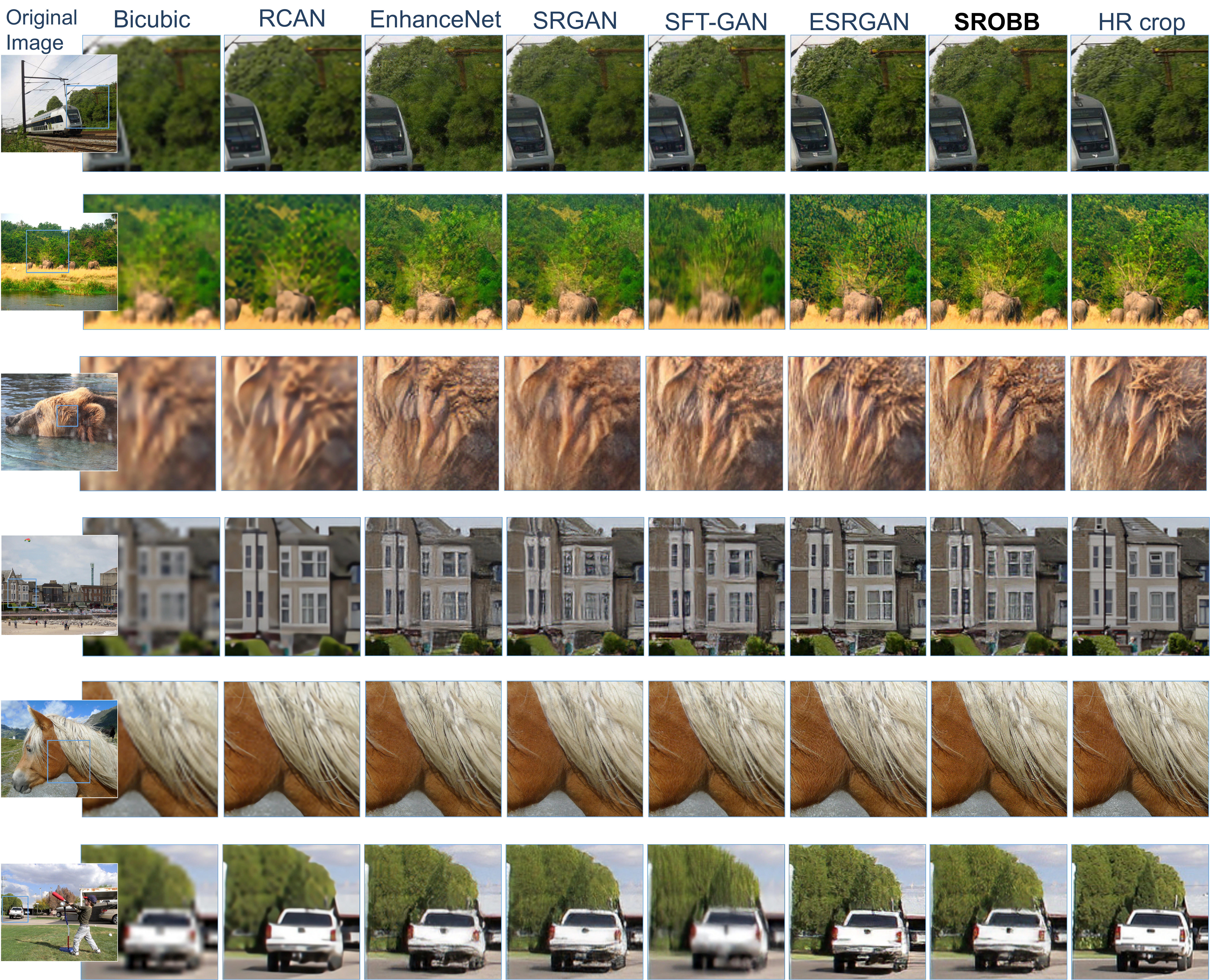}
\end{center}
   \caption{Qualitative results on a subset of the COCO-Stuff dataset \cite{paper_coco_stuff} images. Cropped regions are zoomed in with a factor of 2 to 5 to have a better comparison. Results from left to right: bicubic, RCAN \cite{paper_rcan}, EnhanceNet \cite{paper_enhanced}, SRGAN~\cite{paper_twitter_0}, SFT-GAN~\cite{paper_segmentation}, ESRGAN~\cite{paper_esrgan}, SROBB (ours) and a high resolution image. Zoom in for the best view.}
\label{fig:results_0}
\vspace{-1mm}
\end{figure*}

We randomly chose a set of test images from the COCO-Stuff dataset~\cite{paper_coco_stuff}. In order to have a fair comparison, we re-trained the SFT-GAN\cite{paper_segmentation}, ESRGAN~\cite{paper_esrgan} and SRGAN \cite{paper_twitter_0} methods on the same dataset with the same parameters as ours. For the EnhanceNet and RCAN, we used their pre-trained models by~\cite{paper_enhanced} and \cite{paper_rcan}, respectively. The MATLAB imresize function with a bicubic kernel is used to produce bicubic images. As illustrated in Figure~\ref{fig:results_0}, our method generates more realistic and natural textures by benefiting from our proposed targeted perceptual loss. Although ESRGAN produces very competitive results, it seems that their method is biased towards over-sharpened edges, which sometime leads to an unrealistic reconstruction and dissimilar to ground-truth. 

\subsection{Quantitative Results}
\subsubsection{SSIM, PSNR and LPIPS}
As it is shown in \cite{paper_twitter_0,paper_enhanced,paper_segmentation,paper_report_pirm}, distortion metrics such as the Structural Similarity Index (SSIM)~\cite{1284395} or the Peak Signal to Noise Ratio (PSNR) used as quantitative measurements, are not directly correlated to the perceptual quality; they demonstrate that GAN-based super-resolved images could have higher errors in terms of the PSNR and SSIM metrics, but still generate more appealing images. 

\begin{table}[h]
\begin{center}
\begin{tabular}{ l | c | c c c c}
\Xhline{4\arrayrulewidth}
\small \\[-1em]
\small Image & \small Metric &\small Bicubic & \small LapSRN & \small SRGAN & \small SROBB \\
\hline \\[-1em]
\small \space & \small SSIM & 0.936 & \bf{0.951} & 0.899 & 0.905\\ 
\small baby & \small PSNR & 30.419 & \bf{32.019} & 28.413 & 28.869 \\
\small \space & \small LPIPS & 0.305 & 0.237 & 0.112 & \bf{0.104} \\
\hline \\[-1em]
\small \space & \small SSIM & 0.645 & \bf{0.677} & 0.615 & 0.607\\
\small baboon & \small PSNR & 20.277 & \bf{20.622} & 19.147 & 18.660\\
\small \space & \small LPIPS & 0.632 & 0.537 & \bf{0.220} & 0.245 \\
\Xhline{2\arrayrulewidth}
\end{tabular}
\end{center}
\caption{
Comparison of bicubic interpolation, LapSRN~\cite{paper_lapsrn}, SRGAN~\cite{paper_twitter_0} and SROBB (ours) for the ``baby'' and ``baboon'' images from Set5 and Set14 test sets. Best measures (SSIM, PSNR [dB], LPIPS) are highlighted in bold. The visual comparison is shown in Figure~\ref{fig:set5set14}.}
\label{tab:psnr_ssim}
\vspace{-3mm}
\end{table}

In addition, we used the perceptual similarity distance between the ground-truth and super-resolved images. The Learned Perceptual Image Patch Similarity (LPIPS) metric~\cite{zhang2018} is a recently introduced as a reference-based image quality assessment metric, which seeks to estimate the perceptual similarity between two images. This metric uses linearly calibrated off-the-shelf deep classification networks trained on the very large Berkeley-Adobe Perceptual Patch Similarity (BAPPS) dataset~\cite{zhang2018}, including human perceptual judgments. However, as \cite{gondal2018} also emphasizes, LPIPS has similar trend as distortion-based metrics, e.g., SSIM, and would not necessarily imply photorealistic images.\\
Table~\ref{tab:psnr_ssim} shows the SSIM, PSNR, and LPIPS values estimated between super-resolved images of the ``baby'' and ``baboon'' and their HR counterparts, using bicubic interpolation, LapSRN~\cite{paper_lapsrn}, SRGAN~\cite{paper_twitter_0}, and our method, respectively. Considering this table and the visual comparison of these images in Figure~\ref{fig:set5set14}, we can infer that these metrics would not reflect superior reconstruction quality. Therefore, in the following section, we focus on the user study as the quantitative evaluation.
\subsubsection{User study}
\begin{figure}[h]
\vspace{-1mm}
\begin{center}
   \includegraphics[width=1.0\linewidth]{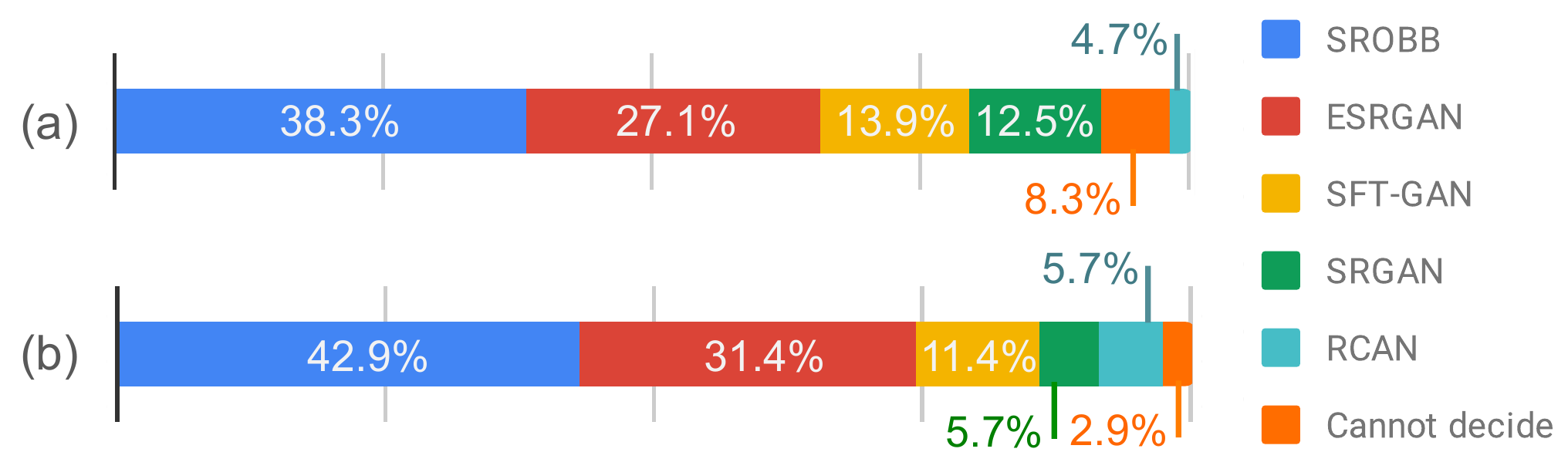}
\end{center}
   \caption{The results of the user study, comparing SROBB (ours) with RCAN~\cite{paper_rcan}, SRGAN~\cite{paper_twitter_0}, ESRGAN~\cite{paper_esrgan} and SFT-GAN~\cite{paper_segmentation} methods. Our method produces visual results that are the preferred choice for the users by a large margin in terms of: (a) percentage of votes, (b) percentage of winning images by majority of votes.}
\label{fig:user_study}
\vspace{-2mm}
\end{figure}

We performed a user study to compare the reconstruction quality of different approaches to see which images are more appealing to users. Five methods were used in the study: 1- RCAN \cite{paper_rcan}, 2- SRGAN \cite{paper_twitter_0}, 3- SFT-GAN~\cite{paper_segmentation}, 4- ESRGAN~\cite{paper_esrgan} and 5- SROBBB (ours).
During the experiment, high-resolution images as well as their five reconstructed counterparts obtained by the mentioned approaches were shown to each user. Users were requested to vote for more appealing images with respect to the ground-truth image. In order to avoid random guesses in case of similar qualities, a choice as ``Cannot decide'' was also designed. Since SFT-GAN uses a segmentation network trained on outdoor categories, for a fair comparison with \cite{paper_segmentation}, we also used 35 images from COCO-Stuff \cite{paper_coco_stuff}, dedicated to outdoor scenes. All images were presented in a randomized fashion to each person. In order to maximize the number of participants, we created our online assessment tool for this purpose.
In total, 46 persons participated in the survey. Figure~\ref{fig:user_study} illustrates that the images reconstructed by our approach are more appealing to the users by a large margin. In terms of number of votes per method, reconstructions by the SROBB got 617 votes, while ESRGAN, SFT-GAN, SRGAN and RCAN methods got 436, 223, 201 and 33 votes, respectively. In addition, the ``Cannot decide'' choice provided in the survey was chosen 100 times. In terms of the best images by majority of votes, among 35 images, SROBB was a dominant choice in 15 images. These results confirm that our approach reconstructs visually more convincing images compared to mentioned methods for the users. Moreover, unlike SFT-GAN, the proposed approach do not require a segmentation map during the test time, while it takes advantage of semantic information and produces competitive results.

\subsubsection{Ablation study}
\begin{figure}[h]
\vspace{-1.5mm}
\begin{center}
   \includegraphics[width=1.0\linewidth]{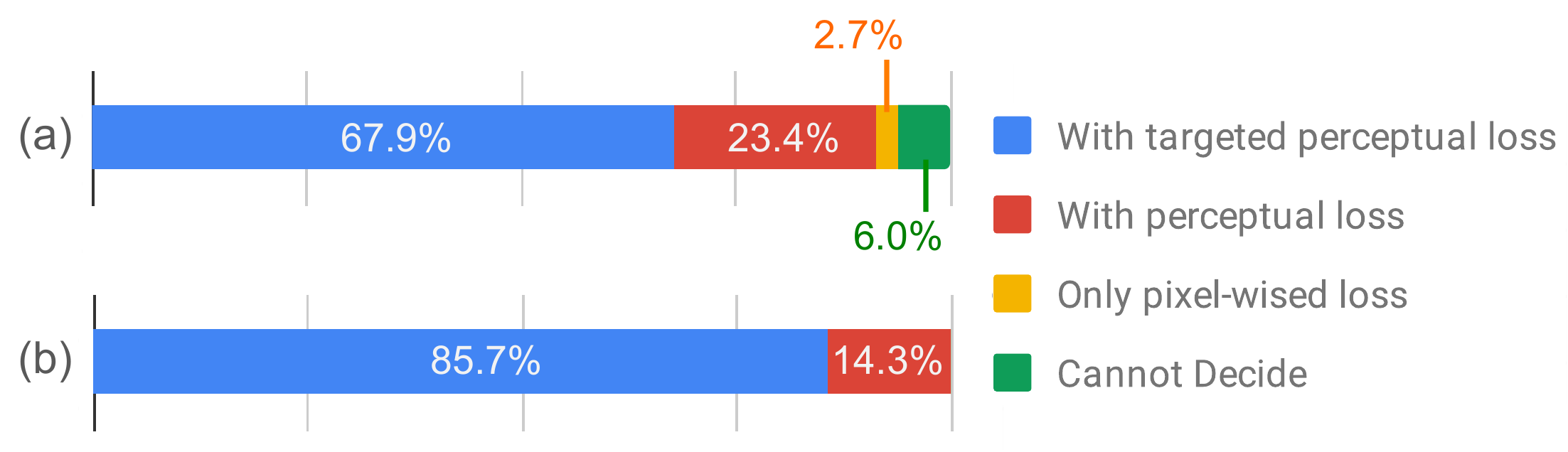}
\end{center}
   \caption{The results of the ablation study showing the effect of the targeted perceptual loss; more convincing results have been obtained by a large margin, in terms of: (a) percentage of votes, (b) percentage of winning images by majority of the votes.}
\label{fig:user_study_ab}
\vspace{-5mm}
\end{figure}

To better investigate the effectiveness of the proposed targeted perceptual loss, we performed a second user study  with similar conditions and procedure to the one in the previous section. Specifically, we study the effect of our proposed targeted perceptual loss; we train our decoder with three different objective functions: 1- pixel-wise MSE only; 2- pixel-wise loss and standard perceptual loss similar to \cite{paper_twitter_0}; and 3- Pixel-wise loss and our proposed targeted perceptual loss (SROBB). The adversarial loss term is also used for both 2 and 3. In total, 51 persons participated in our ablation study survey. Figure~\ref{fig:user_study_ab} shows that users are more convinced when the targeted perceptual loss is used instead of the commonly used perceptual loss. It got 1212 votes, while objective functions 1 and 2 got 49 and 417 votes, respectively. In addition, the ``Cannot decide'' choice was chosen 107 times. In terms of the best images by majority of votes, among 35 images, third objective function was a dominant choice in 30, while 1 and 2 won only in 5 images. Images reconstructed only by the pixel-wise loss had minority number of votes, however, they got considerable number of votes for images in which the ``sky'' was the main class. This can be explained by the over-smoothed nature of the clouds, which suits distortion-based metrics.

%
\subsection{Inference time}
Unlike existing approaches for content-aware SR, our method does not require any semantic information at the input. Therefore, no additional computation is needed at the test time. We reach an inference time of 31.2 frame per second, with a standard XGA output resolution ($1024 \times 768$ in pixels) on a single GeForce GTX 1080 Ti.

\section{Conclusion}
In this paper, we introduced a novel targeted perceptual loss function for the CNN-based single image super-resolution. The proposed objective function penalizes different regions of an image with the relevant loss terms, meaning that using edges' loss for the edges and textures' loss for textures during the training process. In addition, we introduce our OBB labels, created from pixel-wise segmentation label, to provide a better spatial control of the semantic information for the images. This allows our targeted perceptual loss to focus on the semantic regions of an image. Experimental results verify that training with proposed targeted perceptual loss yields perceptually more pleasing results, and outperforms the state-of-the-art SR methods.

{\small
\bibliographystyle{ieee_fullname}
\bibliography{egbib}
}

\end{document}